\documentclass[anon]{nesy2026} 

\usepackage{longtable}

\DeclareMathOperator*{\argsort}{arg\,sort}

\usepackage{booktabs}
\usepackage[T1]{fontenc}
\usepackage[para]{threeparttable}
\usepackage{wrapfig}
\usepackage{algorithm}
\usepackage{algpseudocode}
\usepackage{enumitem}
\usepackage[load-configurations=version-1]{siunitx} 
\usepackage{ml-defs}

\theorembodyfont{\upshape}
\theoremheaderfont{\scshape}
\theorempostheader{:}
\theoremsep{\newline}

\title{G-RRM: Guiding Symbolic Solvers with Recurrent Reasoning Models}

\author{%
  \Name{Timo Bertram\nametag{$^{1}$}}\\
  \Name{Sidhant Bhavnani\nametag{$^{2}$}}\\
  \Name{Richard Freinschlag\nametag{$^{1}$}}\\
  \Name{Erich Kobler\nametag{$^{1,3}$}}\\
  \Name{Andreas Mayr\nametag{$^{1}$}}\\
  \Name{Günter Klambauer\nametag{$^{1,3}$}}\\
  \addr $^{1}$ ELLIS Unit Linz, LIT AI Lab \& Institute for Machine Learning, Johannes Kepler University Linz, Austria\\
  $^{2}$Institute for Symbolic Artificial Intelligence, Johannes Kepler University Linz, Austria\\
  $^{3}$Clinical Research Institute for Medical AI, Johannes Kepler University Linz, Austria
}

\definecolor{mydarkergreen}{rgb}{0.0, 0.4, 0.0}

\begin{document}

\maketitle

\begin{abstract}
Looped Transformers have recently gained significant attention in machine learning as a powerful attention-based architectural paradigm integrating iterative refinement with recurrent parameter sharing. Variants of this architecture have shown promising results on ARC-AGI and Sudoku, both of which are considered to require multi-step reasoning.  Looped transformers give rise to recurrent reasoning models (RRMs), such as the Hierarchical Reasoning Model (HRM) and the Tiny Recursive Model (TRM). Despite their success, RRMs provide no guarantees of correctness: each iteration predicts a complete problem assignment and greedy decoding can yield constraint-violating solutions.
In this work, we focus on SE-RRMs, a symbol-equivariant instantiation of RRMs that exhibits improved extrapolation to larger problem sizes. We propose a neuro-symbolic approach, ``Guiding with Recurrent Reasoning Models'' (G-RRM), which integrates SE-RRMs with symbolic solvers for constraint satisfaction problems. SE-RRMs act as neural solvers that generate full solution proposals and guide classical symbolic solvers, such as backtracking or SAT-based methods like Glucose 4.1 and CaDiCaL 3.0.0, that produce globally correct solutions. Centrally, we investigate when neural guidance with G-RRM improves the search efficiency of symbolic solvers.
Our experiments show that the efficacy of G-RRM depends on two conditions: first, the problem instances must have an expansive combinatorial search space to expose potential gains, and second, the solver architecture must be capable of dynamically overwriting its branching choices to recover when neural hints are imperfect. 
When these conditions hold, guidance drives median conflict counts to zero and yields significant wall-clock speedups: on $9\times9$ Sudoku, where the SE-RRM correctly solves $91.1\%$ of instances, backtracking accelerates by $33.3\times$ and Glucose 4.1 by $1.70\times$ (median, $p<0.001$), with Glucose 4.1 retaining a $1.17\times$ speedup on perfect-hint $25\times25$ grids.
In contrast, CaDiCaL 3.0.0, whose runtime is overhead-dominated and which always respects the injected branching hints rather than overwriting them, shows no significant speedup (median $1.02\times$, n.s.) and even a small significant mean slowdown ($0.90\times$) on $9\times9$. These results delineate the regimes in which neural guidance translates into practical speedups.
\end{abstract}


\begin{figure}[htbp]
\floatconts
  {fig:comparison_NBRRM}
  {\caption{Comparison of a naive search approach (left) to G-RRM (right). Naive search follows a heuristic order of digits, while guided search uses a neural network to order symbol assignments based on confidence.}}
  {\includegraphics[width=0.8\linewidth]{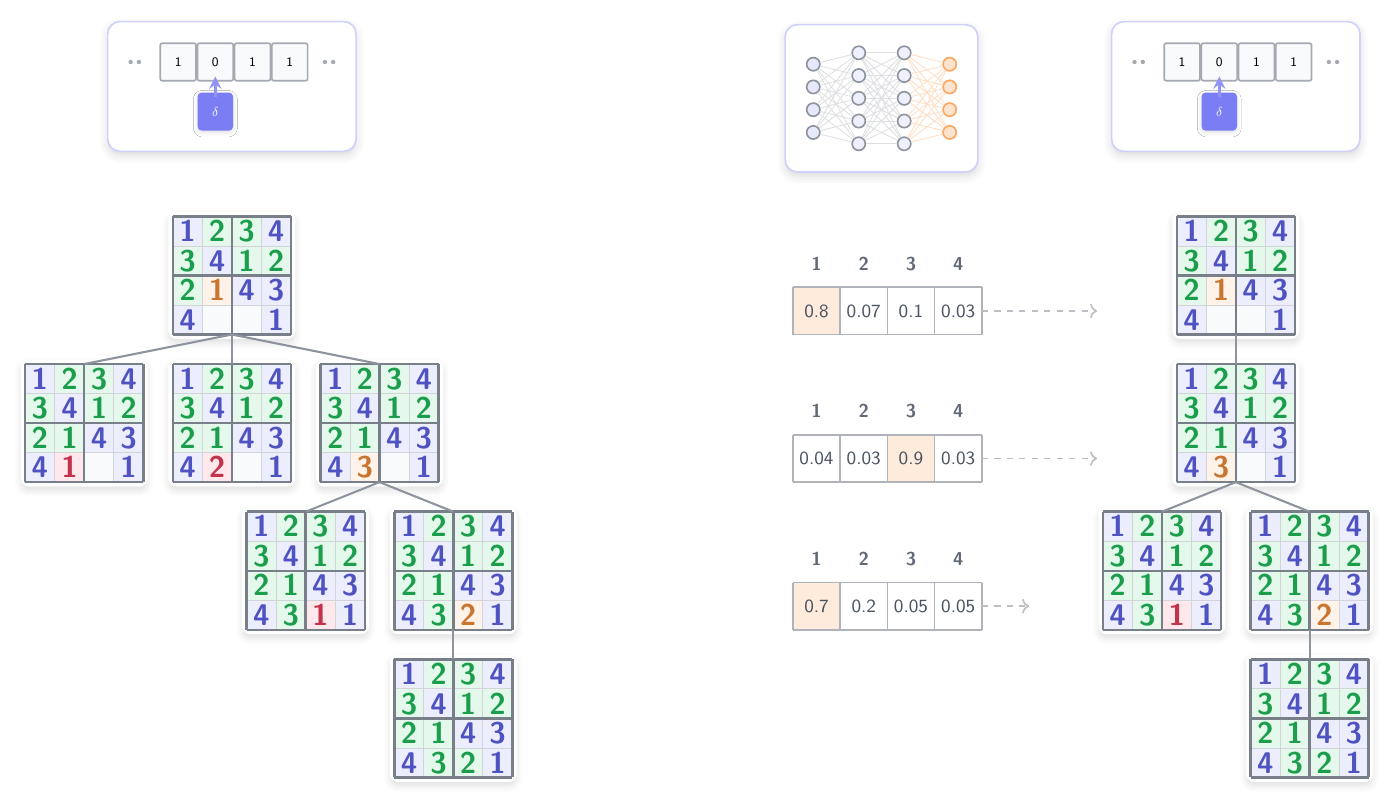}}
\end{figure}

\section{Introduction}
\label{sec:intro}

\paragraph{Looped Transformers/Recurrent Reasoning Models to Solve Combinatorial Problems.}

Due to their parameter efficiency and inherent capability to execute iterative algorithmic procedures through recursive state updates, looped transformers \citep{Liu2024} offer a promising framework for solving complex combinatorial problems. However, they lack the capacity for rigorous logical verification \citep{Renkhoff2024}, raising the question of how to effectively pair their predictive strength with the exact constraints of symbolic methods. At this interface, Recurrent Reasoning Models (RRMs), such as Hierarchical Reasoning Models (HRMs) \citep{wang2025hierarchical} and Tiny Recursive Models (TRMs) \citep{jolicoeur2025less}, have emerged as specialized looped transformers. Despite being called reasoning models, these models do not alter the inherently inductive nature of machine learning, as they perform no deductive logic. Rather, they apply a learned operator to iteratively update an internal state. Nevertheless, they exhibit remarkable predictive performance on heavily constrained tasks. This high accuracy is unlocked by a scalable training methodology that leverages intermediate supervision and stop-gradient operations between individual loop iterations to efficiently optimize deeply looped structures. Consequently, this training paradigm has allowed RRMs to achieve strong empirical results across combinatorial domains, such as Sudoku puzzles and abstract visual reasoning tasks from the ARC-AGI benchmark \citep{chollet2025arc}.


\paragraph{Boolean Satisfiability and Modern SAT Solvers.} Combinatorial problems are traditionally encoded as instances of the Boolean Satisfiability (SAT) problem, the foundational NP-complete problem in theoretical computer science, to provide logical guarantees of correctness. Over the past decades, practical SAT solving has evolved from a purely theoretical challenge into an industrial-grade methodology driven by various advancements in algorithms, heuristics and data structures that can solve large instances.


\paragraph{Sudoku as an Established Combinatorial Benchmark.}
Sudoku puzzles serve as an ideal and widely adopted combinatorial benchmark. Despite their conceptual simplicity, a Sudoku puzzle embodies a rigorous Constraint Satisfaction Problem (CSP) governed by strict, non-local rules. 
Given a grid partially filled with digits, the task is to assign symbols to all remaining cells while ensuring that each row, column, and subgrid contains every symbol exactly once. The generalized decision problem is NP-complete \citep{yato2003complexity}, highlighting its intrinsic computational difficulty. Despite its simple rules, Sudoku poses significant challenges for neural networks. Purely neural approaches often generate locally plausible but globally inconsistent assignments and struggle to extrapolate to unseen grid sizes or symbol sets \citep{long2023large,seely2025sudoku,giannoulis2026teachingtransformers, vamsi2021deep}. Our approach aims to solve this problem by combining neural approaches with exact solvers, which prevents inconsistent assignments and guarantees correctness.

\paragraph{Neurosymbolic Integration.}
To formalize the interplay between these empirical neural approximations and exact deductive logic, we contextualize our approach within Kautz's taxonomy of neural-symbolic integration \citep{kautz2022third}. Architecturally, a pipeline that feeds the predictive outputs of a recurrent reasoning model (RRM) into a downstream symbolic SAT solver most closely aligns with a \textit{Neuro | Symbolic} system (Type 3). While Kautz frames a fully integrated \textit{Neuro[Symbolic]} architecture (Type 6), where symbolic constraints are natively embedded within the neural execution graph, as the ultimate paradigm for combinatorial reasoning, achieving this state with 100\% correctness remains open.

Existing neuro-symbolic frameworks highlight these limitations when applied to hard constraint tasks. For instance, the neuro-symbolic solver of \citet{hathidara2023neuro} remains restricted to Sudoku puzzles with at most eight missing entries, while oscillatory neural network approaches fail when more than roughly half of the cells are unfilled \citep{haverkort2025solving}. Even more general neuro-symbolic learning frameworks, such as semantic constraint strengthening, achieve only modest solve rates of around 28\% and do not reliably solve full, high-difficulty instances \citep{ahmed2023semantic, wang2019satnet}. 

In contrast, recent architectural innovations have closed this empirical performance gap. Recurrent reasoning models have substantially advanced pure neural Sudoku solving, achieving solve rates of roughly 70\% on extremely difficult $9\times9$ puzzles with more than half of the digits missing. Building on this progress, symbol-equivariant RRMs (SE-RRMs) further raise this fully solved rate to over 90\%, while remaining exceptionally compact and trainable from small instances \citep{freinschlag2026symbol}. 


A recurrent reasoning model can propose a full solution to a puzzle in a single forward pass, but it cannot certify it satisfies all constraints. A symbolic solver offers the counterpart: it guarantees correctness without prior knowledge of promising solutions. Coupling the sub-symbolic and symbolic components lets the RRM serve as a prior for the solver's search. When the network is correct, the solver verifies an assignment with no extra search; when it is wrong, the solver falls back on its own completeness guarantees and potentially takes advantage of promising subareas of the search space. 

We show that the benefit of this combination is dependent on two factors: 
(i) the problem's expansive search space
dominates the solver's runtime for an instance, and 
(ii) the ability of the solver to dynamically overwrite its 
branching choices to exploit the hints of the neural model. Consequently, the accuracy of the hints also determines the benefit.

\paragraph{Overview.}
In this work, we introduce Guiding with Recurrent Reasoning Models (G-RRM), a neuro-symbolic framework in which symbolic solvers are guided by the predictive symbol probabilities of a trained Symbol-Equivariant RRM. The paper is organized as follows. Section 2 formalizes the problem setting and defines the search-guidance mechanism for backtracking and CDCL solvers. Section 3 presents our experimental evaluation across Sudoku grids of varying sizes, benchmarking G-RRM against naive backtracking and the modern SAT solvers Glucose 4.1 and CaDiCaL 3.0.0. Section 4 concludes our findings and identifies the conditions under which neural guidance improves solver efficiency. Finally, Section 5 outlines limitations of our work and Section 6 describes future work.

\section{Problem Setting and Methodological Background}

\subsection{Recurrent Reasoning Models}
\label{sec:background_rrm1}
We briefly review recurrent reasoning models (RRMs), including HRM \citep{wang2025hierarchical}, TRM \citep{jolicoeur2025less}, and SE‑RRM \citep{freinschlag2026symbol}. Throughout, we follow the notation of \citet{freinschlag2026symbol}. For a more in-depth explanation, please refer to \appendixref{app:rrm}.

\paragraph{Problem Setting and Notation.}
\label{sec:background_notation}
RRMs have been developed for structured problem‑solving tasks, such as Sudoku solving or maze navigation, where both the input and the possible solutions consist of assignments of discrete values to grid cells.
A task instance is represented as a tuple of discrete symbols
$\BX = (x_1,\dots,x_I), \; x_i \in \Sigma,$
where $I$ is the number of grid cells 
and $\Sigma$ is a symbol alphabet with
$K := |\Sigma|$ symbols (e.g., digits 
and an additional mask symbol). 
A solution is represented analogously as 
$\BY = (y_1,\dots,y_I)$ with $y_i \in \Sigma$.

\paragraph{Operational Principle.}
\label{sec:background_hrm_trm}
RRMs are specialized looped transformer architectures proposed for multi-step reasoning, where the same Transformer-based computational block is repeatedly reused to iteratively refine an internal latent representation (see \eqref{eq:bg_rrm_fp} in \appendixref{app:rrm}). As looped transformers, RRMs are highly parameter-efficient due to extensive parameter sharing across recurrent iterations. Their practical success is largely attributed to a scalable training methodology based on deep supervision and stop-gradient operations. HRM and TRM mainly differ in how they architecturally realize recurrent computation.

\paragraph{Symbol-Equivariant RRMs (SE-RRMs).}
\label{sec:background_serrm}

SE-RRMs modify the standard RRM formulation to achieve symbol equivariance. They introduce an explicit symbol axis into the latent representation, yielding rank-3 tensors of shape $D\times I\times K$ instead of the standard rank-2 representations of shape $D\times I$, where $D$ is the feature dimension. This change is reflected in corresponding adaptations of both the input encoding and the recurrent update, in particular through the inclusion of attention over the symbol dimension. 

SE-RRM provides per-cell symbol distributions that remain meaningful even when the symbol set changes, making it a natural candidate as a learned policy for symbolic search, e.g., for variable ordering in backtracking. This is the key interface for guiding symbolic methods.

\subsection{Symbolic Search and Verification}

\subsubsection{Backtracking Search}
\label{sec:backtracking}

Backtracking uses depth-first search to explore possible solutions in a search tree while continuously checking constraints and undoing choices when violations occur. For Sudoku, it starts from a partially filled puzzle and repeatedly selects empty cells to fill. Nodes in the search tree correspond to partially filled puzzles, while edges represent valid digit assignments to a previously empty cell.


Whenever the algorithm reaches a state in which an empty cell has no valid digit available, the current partial solution cannot be extended to a complete solution. In this case, the algorithm backtracks by undoing the most recent assignment and exploring alternative edges. This process is repeated until either a complete valid Sudoku solution is found or all possible assignments have been exhausted.

This strategy guarantees completeness: if a valid Sudoku solution exists, backtracking will eventually find it by systematically exploring all feasible assignments.

\subsubsection{Sudoku as a SAT Problem}

To solve Sudoku using highly optimized modern SAT solvers, the puzzle must be encoded as a formula in Conjunctive Normal Form (CNF). We introduce Boolean variables $x_{r,c,v}$ indicating that cell $(r,c)$ contains value $v$, where $r,c,v \in \{1,\dots,N\}$ for an $N \times N$ grid. Every Sudoku rule constrains a given region (a cell, row, column, or $\sqrt{N}\times\sqrt{N}$ box) to contain exactly one instance of a value. We enforce this using a standard structural encoding where an \emph{At-Least-One} (ALO) clause $\bigvee_{v=1}^{N} x_{r,c,v}$ ensures that every cell is assigned at least one digit, while pairwise \emph{At-Most-One} (AMO) clauses $(\lnot x_{r,c,v} \lor \lnot x_{r,c,w})$ for $v \neq w$ forbid duplicate assignments so no cell contains multiple values. This identical ALO/AMO pattern is applied across all rows, columns, and boxes to enforce global constraint satisfaction. Prefilled cells are permanently fixed to true via unit clauses $x_{r,c,v}$.

The resulting CNF formula is satisfiable if and only if a valid Sudoku completion exists. The full encoding requires $N^3$ variables and yields $11{,}988$ base clauses for $9\times9$, $123{,}904$ for $16\times16$, and $752{,}500$ for $25\times25$ grids.

\subsubsection{SAT Solving}

Modern Conflict-Driven Clause Learning (CDCL) SAT solvers, like Glucose 4.1 (\texttt{glucose4}) \citep{glucose} and CaDiCaL 3.0.0 (\texttt{cadical3) }\citep{cadical}, evaluate propositional formulas in CNF by combining systematic search with logical deduction and dynamic clause database management. Rather than traversing the search space chronologically, a CDCL solver relies on three key foundational features, which we discuss below.

\paragraph{Heuristic Selection (Variables and Phases).} The solver dynamically selects unassigned variables using scoring metrics like Variable State Independent Decaying Sum (VSIDS) \citep{chaff}. For each chosen variable, the solver must also select a literal polarity (phase) to determine which truth value to explore first.
\paragraph{Conflict-Driven Learning and Backjumping.} When an assignment triggers a logical contradiction, the solver analyzes the root cause via an implication graph to derive a globally valid \textit{learned clause}. It then executes a non-chronological \textit{backjump} to the deepest decision level where the learned clause becomes a unit, permanently pruning the invalid search space.
\paragraph{Restarts and Database Management.} To escape local minima and heavy-tailed runtime distributions, the solver periodically flushes its active assignment stack (the trail) and returns to the root level. Crucially, it preserves its cached variable phases, heuristic scores, and highly active learned clauses during a restart. To prevent memory bloat, it concurrently purges low-utility learned clauses.

\subsection{G-RRM: Guiding Symbolic Solvers with Recurrent Reasoning Models}
\label{sec:g_rrm}

\subsubsection{General Strategy}

We propose G-RRM, a framework that couples recurrent reasoning models with symbolic solvers through search guidance (Figure~\ref{fig:comparison_NBRRM}). We instantiate the recurrent model as an SE-RRM, trained on pairs of partially filled Sudoku grids and their corresponding solutions.

At inference time, SE-RRM assigns a score to each variable–value pair, yielding a matrix $\hat{\mathbf Y}\in\mathbb{R}^{I\times K}$, where $\hat{\mathbf Y}_{i,d}$ reflects the model’s preference for assigning value $d$ to variable $i$. For each variable $i$, these scores define a permutation
\begin{equation}
\pi_i = \argsort_{d\in \{{1,\dots,K}\}} \hat{\mathbf Y}_{i,d},
\end{equation}
where $\argsort$ returns indices in ascending order of the scores, i.e., from lowest to highest value. The most preferred value for a variable $i$ is therefore given by the last element of the permutation, $d_i^* = \pi_i(K)$.

The resulting ordering guides a downstream symbolic solver during search by prioritizing candidate values in decreasing order of their SE-RRM scores, thereby affecting variable selection, branching decisions, or the order of candidate exploration.

When SE-RRM predictions are accurate, the solver can reach a solution with little or no additional search, effectively reducing the process to a sequence of decisions that can be verified efficiently. When predictions are imperfect, they still bias the search towards promising regions of the search space, while the solver explores alternatives as needed.

Importantly, G-RRM does not restrict the search space; it only determines the order of exploration while preserving all feasible assignments. Consequently, correctness and completeness of the underlying symbolic solver remain unchanged, and improvements arise solely from more efficient search ordering.

Because neural guidance influences only the direction of branching decisions, its efficacy depends on both the instance and the solver architecture. G-RRM can yield substantial benefits on search-dominated instances where decision-tree exploration forms the primary bottleneck, such as in backtracking and \texttt{glucose4}. However, these gains can be masked in solvers with high structural or startup overhead, such as \texttt{cadical3}, where initial startup costs persist independently of conflict reduction. 

\subsubsection{G-RRM for Backtracking}

The efficiency of backtracking strongly depends on the order in which cells and digits are explored. Good exploration directly reduces the size of the search tree and leads to faster solution times. Our approach combines a classical cell-selection heuristic with neural guidance for digit selection. We use backtracking as the simplest example of G-RRM, making the improvements more pronounced, as the efficiency of backtracking is directly tied to the explored search space.

\paragraph{Cell Selection Heuristic.}
At each step, the algorithm must choose one of the remaining empty cells. To reduce the size of the search tree, we select the cell with the fewest valid candidate digits. This is the well-known minimum remaining values heuristic and helps expose contradictions early, often reducing the amount of search required to solve the puzzle.

\paragraph{Digit Selection through Neural Guidance.}
After selecting a cell, the algorithm must decide in which order its valid candidate digits should be explored. Classical backtracking implementations typically use a fixed ordering. Instead, we use the predictions of a trained SE-RRM to guide this decision.

\subsubsection{G-RRM for CDCL SAT Solvers}

Neural guidance in CDCL solvers attaches to the phase selection: at each decision point, the solver chooses an unassigned variable together with a polarity (True or False) to try first. This choice determines which region of the search space the solver explores before any conflict is encountered. Correct phase assignment eliminates backtracking entirely; wrong assignment forces the solver to discover and escape an infeasible subspace before recovering. We exploit solver control over this heuristic by initializing phases from SE-RRM predictions before search begins by setting variable phases from the highest-scored digit $d_i^*$.  The Boolean variable $x_{r_i,c_i,d_i^*}$ is initialized to 1, and the variables $x_{r_i,c_i,d}$ are initialized as 0 for all $d\neq d_i^*$. Because SE-RRM suggests full solution proposals over all grid cells simultaneously, we map the network's high-confidence predictions directly onto the initial Boolean polarities (phases) of the CNF variables. When the model's hints are highly accurate, the solver can traverse a valid path to the solution with zero or minimal conflicts. The normal configuration omits this step, relying solely on the solver's built-in heuristics.

We leverage two popular CDCL solvers to evaluate how they interact with this neural phase guidance: \texttt{glucose4} and \texttt{cadical3}. Crucially, these solvers manage phase initialization and internal recovery differently when external hints conflict with the underlying logic:

\begin{itemize}
\item \texttt{cadical3}: This solver strictly honors externally set literal phases. Provided phases persist across restarts, forcing the solver to consistently prioritize the externally injected search direction unless explicitly overridden by hard learned clause constraints.
\item \texttt{glucose4}: This solver adopts a more fluid approach and may override initial external phases on restart via its native VSIDS-driven reinitialization mechanism. This design yields a significantly faster self-correction loop, allowing the solver to rapidly recover and diverge from faulty hints directing the search into an infeasible subspace.
\end{itemize}

\section{Experimental Evaluation}

In this section, we evaluate G-RRM by benchmarking different solvers’ default configurations against their SE-RRM–guided variants to quantify the impact of neural guidance on solver efficiency for Sudoku puzzles of different sizes. For training details of the SE-RRM, see \appendixref{app:trDetails}. Additional CP-SAT experiments are located in \appendixref{app:cpsat}.

Because reduced conflicts directly lead to runtime savings for search‑dominated solvers, we quantify the efficiency of G-RRM via changes in conflict counts alongside wall‑clock time (see \tableref{tab:sat_conflicts_full} and \tableref{tab:sat_timing_full}).
In addition, we report the fully‑solved rate (FSR), which denotes the fraction of Sudoku instances that SE-RRM immediately predicts correctly, for each benchmarked grid size.

For the SAT solvers, we interface with \texttt{glucose4} and \texttt{cadical3} via the \texttt{PySAT} Python library \citep{pysat2018, pysat2024}, integrating the neural outputs directly into the solver via phase initialization. For the wall‑clock time experiments, we separately evaluate the advantage of neural guidance for cases with perfect neural predictions and for cases where the predictions contain errors. We do not aim to directly compare the performance of backtracking and the CDCL solvers, as we use a custom Python implementation of backtracking. Rather, we compare the \textit{normal} and \textit{guided} version of each solver.

\paragraph{Conflict Counts (Table~\ref{tab:sat_conflicts_full}).}

\begin{table}[htbp]
\centering
\caption{Conflict count percentiles (Normal vs.\ Guided) across grid sizes and solvers.
  Bold marks the best value at each percentile. Backtracking conflicts denote DFS
  search dead-ends and are not directly comparable in magnitude to CDCL conflicts.}
\label{tab:sat_conflicts_full}
\setlength{\tabcolsep}{4pt}
\renewcommand{\arraystretch}{1.1}
\begin{threeparttable}
\resizebox{\linewidth}{!}{%
\begin{tabular}{lclccccccc}
\toprule
Grid & SE-RRM FSR & Solver & Mode & p50 & p75 & p90 & p95 & p99 & Max \\
\midrule
\multirow{6}{*}{$9{\times}9$} & \multirow{6}{*}{$91.1\%$\tnote{b}}
& \multirow{2}{*}{\texttt{backtracking}\tnote{a}}
  & Normal & 2865.0 & 9747.8 & 17932.9 & 24791.5 & 42058.6 & \textbf{192185.0} \\
& & & Guided & \textbf{0.0} & \textbf{0.0} & \textbf{0.0} & \textbf{5166.3} & \textbf{18457.5} & 555689.0 \\
\cmidrule{3-10}
& & \multirow{2}{*}{\texttt{glucose4}}
  & Normal & 19.0 & 39.0 & 62.0 & 75.9 & 116.6 & 201 \\
& & & Guided & \textbf{0.0} & \textbf{0.0} & \textbf{0.0} & \textbf{30.8} & \textbf{73.5} & \textbf{188} \\
\cmidrule{3-10}
& & \multirow{2}{*}{\texttt{cadical3}}
  & Normal & 13.0 & 31.0 & 53.0 & 72.0 & 102.0 & 155 \\
& & & Guided & \textbf{0.0} & \textbf{0.0} & \textbf{0.0} & \textbf{27.7} & \textbf{80.3} & \textbf{174} \\
\midrule
\multirow{4}{*}{$16{\times}16$} & \multirow{4}{*}{$22.0\%$}
& \multirow{2}{*}{\texttt{glucose4}}
  & Normal & 93.0 & 173.8 & \textbf{242.6} & \textbf{284.1} & \textbf{507.9} & \textbf{598} \\
& & & Guided & \textbf{43.5} & \textbf{151.3} & 277.0 & 395.8 & 767.1 & 779 \\
\cmidrule{3-10}
& & \multirow{2}{*}{\texttt{cadical3}}
  & Normal & 72.5 & 122.8 & \textbf{184.1} & \textbf{258.1} & \textbf{469.3} & \textbf{502} \\
& & & Guided & \textbf{56.5} & \textbf{129.8} & 257.3 & 366.5 & 477.1 & 581 \\
\midrule
\multirow{4}{*}{$25{\times}25$} & \multirow{4}{*}{$51.1\%$}
& \multirow{2}{*}{\texttt{glucose4}}
  & Normal & 49.5 & 78.3 & 210.8 & \textbf{244.6} & \textbf{392.3} & \textbf{468} \\
& & & Guided & \textbf{0.0} & \textbf{51.0} & \textbf{136.8} & 264.0 & 488.6 & 502 \\
\cmidrule{3-10}
& & \multirow{2}{*}{\texttt{cadical3}}
  & Normal & 28.0 & 71.0 & \textbf{147.3} & 174.9 & 392.2 & 410 \\
& & & Guided & \textbf{0.0} & \textbf{49.8} & 148.7 & \textbf{161.4} & \textbf{328.6} & \textbf{366} \\
\bottomrule
\end{tabular}
}
\begin{tablenotes}
\footnotesize
\item[a] Standard backtracking is too inefficient to solve puzzles larger than $9\times9$.\\
\item[b] In this work, we use a different set of $9\times9$ puzzles than the authors of SE-RRM, which explains the  \\ slightly lower fully-solved rate on $9\times9$ Sudokus compared to \citet{freinschlag2026symbol}.
\end{tablenotes}
\end{threeparttable}
\end{table}

Neural guidance tends to reduce conflict counts most consistently at the median: on $9 \times 9$, it drives the median (and p90) to zero across solvers, reflecting the instances perfectly solved by the SE-RRM. On $16  \times 16$ and $25 \times 25$, it collapses or markedly lowers the median (e.g., $-53.2\%$ for glucose4 and $-22.1\%$ for cadical3 at $16 \times 16$; zero median at $25 \times 25$), with smaller or mixed effects in the upper tail. These conflict reductions do not necessarily translate into significant wall-clock speedups, presumably because runtime is not search-dominated.

\paragraph{Wall-Clock Time (Table~\ref{tab:sat_timing_full}).}

\begin{table}[htbp]
\centering
\caption{Wall-clock solve times (ms) of normal and guided SAT
  solvers across all grid sizes. Speedup > 1 indicates that neural guidance reduces solve time. Significance of each speedup is from a paired test on per-instance solve times across the test set (Wilcoxon signed-rank for Median rows; paired t-test for Mean rows): *** p<0.001, ** p<0.01, * p<0.05, n.s. = not significant. Seeds were collapsed per instance (mean over 3 runs) before aggregation. Note that the significance for the Median rows is more relevant, as the t-test on Mean rows is less informative due to the heavy-tailed distributions.}
\label{tab:sat_timing_full}
\setlength{\tabcolsep}{4pt}
\renewcommand{\arraystretch}{1.1}
\begin{threeparttable} 
\resizebox{\linewidth}{!}{%
\begin{tabular}{lllcccccccccccc}
\toprule
Grid & Solver & Metric
  & \multicolumn{4}{c}{All Puzzles}
  & \multicolumn{4}{c}{Perfect ($\mathrm{acc}=1.0$)}
  & \multicolumn{4}{c}{Imperfect ($\mathrm{acc}<1.0$)} \\
& & & Normal & Guided & Speedup & Significant & Normal & Guided & Speedup & Significant & Normal & Guided & Speedup & Significant \\
\midrule
\multirow{6}{*}{$9{\times}9$}
& \multirow{2}{*}{\texttt{backtracking}\tnote{a}}
  & Median\,(ms) & 20.503 & \textbf{0.617} & 33.251$\!\times$ &*** &18.230 & \textbf{0.606} & 30.101$\!\times$ &***& \textbf{41.535} & 42.601 & 0.975$\!\times$ & n.s. \\
& & Mean\,(ms)   & 47.479 & \textbf{6.133} & 7.741$\!\times$ & *** & 45.556 & \textbf{0.601} & 75.750$\!\times$ & *** & 67.197 & \textbf{62.848} & 1.069$\!\times$ & n.s. \\
\cmidrule{2-15}
& \multirow{2}{*}{\texttt{glucose4}}
 & Median\,(ms)                                       & 0.348          & \textbf{0.205}  & 1.699$\!\times$ & ***     & 0.337         & \textbf{0.202}  & 1.666$\!\times$ & ***     & 0.451          & \textbf{0.450}  & 1.003$\!\times$ & n.s.    \\
                 &                                                    & Mean\,(ms)      & 0.419          & \textbf{0.245}  & 1.710$\!\times$ & *** & 0.404          & \textbf{0.213}  & 1.896$\!\times$ & ***   & 0.575          & \textbf{0.573}  & 1.003$\!\times$ & n.s. \\
            \cmidrule{2-15}
                             
& \multirow{2}{*}{\texttt{cadical3}}                             
                 & Median\,(ms)                                       & 5.517          & \textbf{5.405}  & 1.021$\!\times$ & n.s.     & 5.455         & \textbf{5.307}  & 1.028$\!\times$ & n.s.     & 6.074          & \textbf{5.876}  & 1.034$\!\times$ &  n.s.   \\
                 &                                                    & Mean\,(ms)      & \textbf{4.262} & 4.759           & 0.896$\!\times$ & *** & \textbf{4.125} & 4.622           & 0.892$\!\times$ & ***   & \textbf{5.672} & 6.167           & 0.920$\!\times$ & n.s. \\
            \midrule
            \multirow{4}*{$16{\times}16$}
& \multirow{2}{*}{\texttt{glucose4}}                         
                 & Median\,(ms)                                       & 2.671          & \textbf{1.998}  & 1.337$\!\times$ & n.s.     & 2.967         & \textbf{1.346}  & 2.204$\!\times$ & ***     & 2.647          & \textbf{2.627}  & 1.008$\!\times$ & n.s.    \\
                 &                                                    & Mean\,(ms)      & 3.158          & \textbf{2.974}  & 1.062$\!\times$ & n.s. & 3.245          & \textbf{1.369}  & 2.371$\!\times$ & ***   & \textbf{3.134} & 3.427           & 0.914$\!\times$ & n.s. \\
            \cmidrule{2-15}
                             
& \multirow{2}{*}{\texttt{cadical3}}                               
                 & Median\,(ms)                                       & 22.903         & \textbf{22.599} & 1.013$\!\times$ & n.s.     & 22.767        & \textbf{21.680} & 1.050$\!\times$ & *     & \textbf{22.903} & 22.988          & 0.996$\!\times$ & n.s.    \\
                 &                                                    & Mean\,(ms)      & 22.730         & \textbf{22.457} & 1.012$\!\times$ & n.s. & 22.178         & \textbf{20.935} & 1.059$\!\times$ & n.s.   & 22.886         & \textbf{22.886} & 1.000$\!\times$ & n.s. \\
            \midrule
            \multirow{4}*{$25{\times}25$}
& \multirow{2}{*}{\texttt{glucose4}}                               
                 & Median\,(ms)                                       & 6.403          & \textbf{5.779}  & 1.108$\!\times$ & **     & 6.200         & \textbf{5.305}  & 1.169$\!\times$ & ***     & 6.538          & \textbf{6.494}  & 1.007$\!\times$ & n.s.    \\
                 &                                                    & Mean\,(ms)      & 6.778          & \textbf{6.334}  & 1.070$\!\times$ & * & 6.374          & \textbf{5.415}  & 1.177$\!\times$ & ***   & \textbf{7.201} & 7.294           & 0.987$\!\times$ & n.s. \\
            \cmidrule{2-15}
                             
& \multirow{2}{*}{\texttt{cadical3}}                             
                 & Median\,(ms)                                       & 74.474         & \textbf{73.424} & 1.014$\!\times$ & n.s.     & 72.419        & \textbf{71.739} & 1.009$\!\times$ & n.s.     & \textbf{77.260} & 77.328          & 0.999$\!\times$ & n.s.    \\
                 &                                                    & Mean\,(ms)      & 73.104         & \textbf{71.989} & 1.016$\!\times$ & n.s. & \textbf{69.422} & 70.690          & 0.982$\!\times$ & n.s.   & 76.955         & \textbf{73.346} & 1.049$\!\times$ & n.s. \\
            \bottomrule
        \end{tabular}
}
\begin{tablenotes}
\footnotesize
\item[a] Standard backtracking is too inefficient to solve puzzles larger than $9\times9$.
\end{tablenotes}
\end{threeparttable}
\end{table}

Backtracking shows large, statistically significant speedups on all puzzles and perfect puzzles ($\mathbf{33.251}\!\times$ and $\mathbf{30.101}\!\times$ respectively) in $9\times9$, as reduced conflicts directly lead to faster runtime. On imperfect solutions, the results are mixed and not significant.

\texttt{glucose4} consistently converts conflict savings into statistically significant wall-clock speedups: $\mathbf{1.699}\!\times$
overall and $\mathbf{1.666}\!\times$ on perfect-hint instances at $9{\times}9$, with
perfect-hint speedups of $\mathbf{2.204}\!\times$ and $\mathbf{1.169}\!\times$ at $16{\times}16$
and $25{\times}25$, respectively. Speedups on all puzzles and imperfect hints for $16\times16$ and $25\times25$ are mostly non-significant, only reaching moderate significance on all puzzles for $25\times25$. The diminishing returns at larger scales reflect a growing share of solve time spent outside the search.

\texttt{cadical3} shows no consistent wall-clock benefit and provides the only statistically significant slowdown using G-RRM. 
The solver is overhead-dominated: its internal bookkeeping is largely
independent of conflict count, so eliminating conflicts does not proportionally reduce
runtime. On $9{\times}9$, phase initialization adds ${\sim}1.7$\,ms
on instances \texttt{cadical3} would have solved trivially, producing a
\textbf{$0.896\!\times$} aggregate mean.
On the more reliable median, \texttt{cadical3} shows no significant change ($1.021\!\times$, n.s.); the only significant effect is a small mean slowdown  attributable to fixed phase-initialization overhead.
On $16\times16$ and $25{\times}25$, search accounts for a tiny fraction of the solve budget, leaving guidance as a non-significant, near-zero impact.

\section{Conclusion}

We propose G-RRM, a neuro-symbolic framework in which a symbol-equivariant recurrent reasoning model (SE-RRM) guides the search of exact symbolic solvers. 
The neural model provides probabilistic preferences over candidate assignments from a single forward pass, while a symbolic solver enforces global constraints and guarantees correctness. 
This combination reduces the effective search space compared to only using a solver and still retains the completeness of the symbolic algorithm; improvements arise from more efficient search. In essence, G-RRM converts a high-accuracy but unverified neural solver into one with guaranteed correctness. 

The central question of this work is under what conditions G-RRM reduces search cost. Our results show that reductions in search cost are strongly tied to the solver architecture and the instance-level correctness of the neural model. On perfect-hint instances, guidance resolves puzzles with zero conflicts across all grid sizes and solvers. For the search-dominated backtracking and \texttt{glucose4}, this translates into statistically significant wall-clock speedups in all cases. For the overhead-dominated \texttt{cadical3} it does not, as its runtime is largely independent of conflict count. 
On imperfect-hint instances, the outcome depends on solver architecture and hint quality. Although we see slight speedups in some instances, e.g., a $1.069\!\times$ speedup in mean time for backtracking on $9\times9$, no result is statistically significant. This can be explained by the confidence of the SE-RRM: independent of being correct in its assignments, the majority of predictions strongly favor one option. When the model is confidently incorrect, it leads the search towards incorrect paths, which cancel the speedup of promising directions.

\section{Limitations}
Our evaluation isolates the effect of neural guidance, which introduces some limitations to the results.

First, all reported wall‑clock times refer to the symbolic solver. Inference time for SE‑RRM is not included, as it was precomputed separately. Incorporating inference time would entangle batch size, GPU hardware, and implementation‑specific optimizations, making comparisons difficult. Our objective in this work is to isolate the impact of neural guidance itself, rather than to maximize engineering performance. Accordingly, our SE‑RRM implementation is a straightforward research prototype and does not exploit the full range of possible performance optimizations.

Second, our study is confined to Sudoku. While Sudoku is an established combinatorial benchmark, we do not confirm our results on other domains.

\section{Future Work}
Beyond Sudoku, G-RRM suggests a general strategy for combinatorial problems whose
symbolic search cost grows exponentially with size. We call it
\emph{Solve--Learn--Extrapolate (SLE)}: 
(i) \emph{solve} small instances exactly
with a symbolic solver, 
(ii) \emph{learn} an SE-RRM on these valid solutions,
and (iii) \emph{extrapolate} to larger instances, 
where SE-RRM hints guide the
solver through the search space. 
Since G-RRM preserves
completeness, new solutions are valid and 
can be fed back to
re-train the SE-RRM before advancing to the next size,
lifting the frontier of
tractable sizes using only valid data. 
Our experiments realize a first step of this loop: the SE-RRM trained on
$9\times9$ Sudokus and fine-tuned on a small number of larger grids
(Appendix~\ref{app:trDetails}) produces hints that guide solvers on $16\times16$
and $25\times25$ instances, demonstrating SLE in miniature. 
Assessing SLE on harder problem families is a promising direction for future work.

\section*{Acknowledgements}
The ELLIS Unit Linz, the LIT AI Lab, the Institute for Machine Learning, are supported by the Federal State Upper Austria. We thank the projects FWF AIRI FG 9-N (10.55776/FG9), AI4GreenHeatingGrids (FFG-899943), Stars4Waters (HORIZON-CL6-2021-CLIMATE-01-01), FWF Bilateral Artificial Intelligence (10.55776/COE12). We thank NXAI GmbH, Audi AG, Merck Healthcare KGaA, GLS (Univ. Waterloo), T\"{U}V Holding GmbH, Software Competence Center Hagenberg GmbH, dSPACE GmbH, TRUMPF SE + Co. KG.

\bibliography{bib}

\newpage

\appendix

\section{Recurrent Reasoning Models}
\label{app:rrm}

\begin{figure}[htbp]
\floatconts
  {fig:rrm}
  {\caption{Principle designs of Recurrent Reasoning Models (Source: \citealt{freinschlag2026symbol}). \textbf{Left:} HRM/TRM compared to SE-RRM which introduces an explicit symbol dimension that enables permutation equivariance over symbols and allows for problem size extrapolation. \textbf{Right:} Permutation equivariance explained.}}
  {\includegraphics[width=0.4\linewidth]{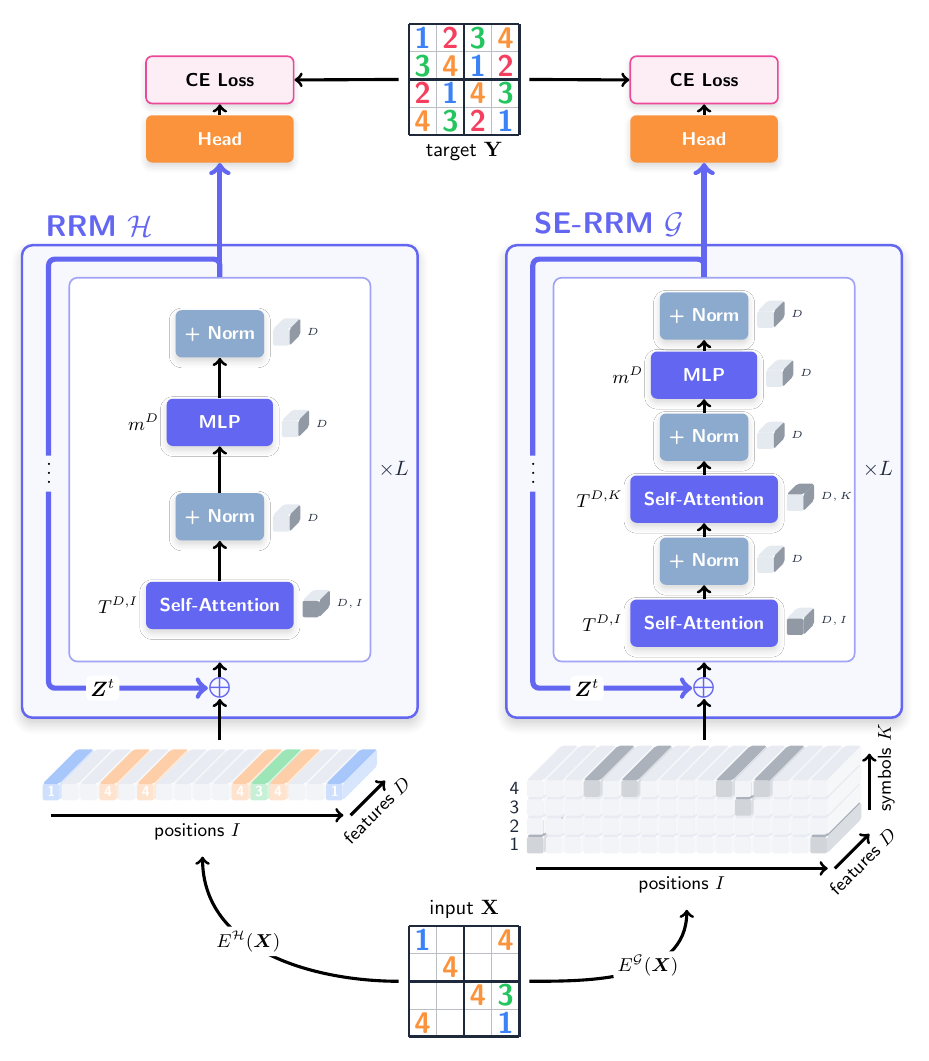}\hspace{0.5cm}
   \includegraphics[width=0.4\linewidth]{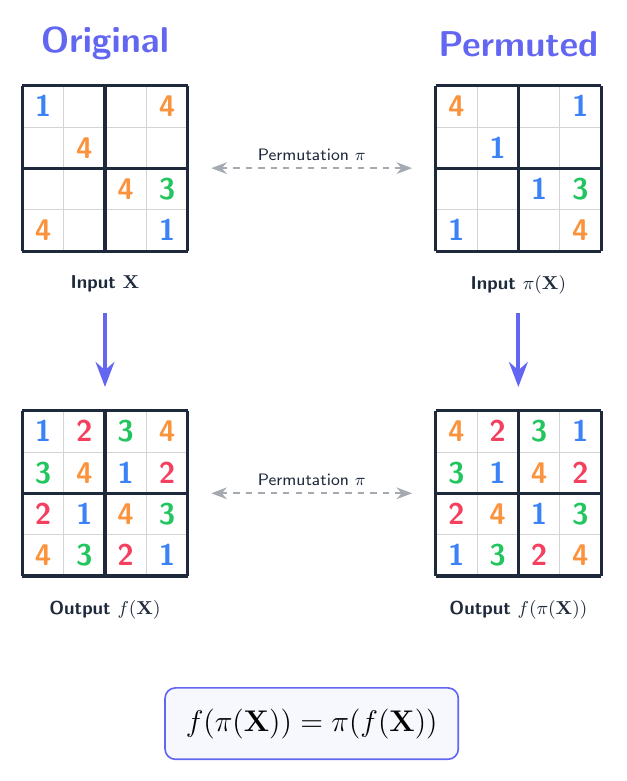}}
\end{figure}

\subsection{Recurrent Reasoning Models: Fixed-Point Iteration with Deep Supervision}
\label{sec:background_rrm2}
Recurrent reasoning models (RRMs) perform iterative refinement via a fixed-point iteration on a recurrent state. An overview of our notation is shown in Table~\ref{tab:notation}.
Let $\BZ^t \in \mathbb{R}^{D\times I}$ denote the recurrent state at iteration $t$, where $D$ is the feature dimension. 
An input embedding $E^{\mathcal H}$ maps the symbolic instance into the same shape, $E^{\mathcal H}(\BX)\in\mathbb{R}^{D\times I},$
typically as a sum of a symbol embedding 
and a positional encoding.

An \emph{RRM block} $\mathcal H$ is a mapping
$\mathcal H:\mathbb{R}^{D\times I}\times \mathbb{R}^{D\times I}\to \mathbb{R}^{D\times I},$
which updates the state $\BZ$ as
\begin{align} \label{eq:bg_rrm_fp}
\BZ^{t+1} = \mathcal H \left( E^{\mathcal H}(\BX),\BZ^t \right),
\end{align}
where $\BZ^0$ is a learned initial state.

Variants of the RRM block may omit the input embedding at some steps or 
inject a previous recurrent state $\BZ^{t-r_{prev}}$ (with $r_{prev}$ being a hyperparameter). We capture this by
\begin{align} \label{eq:bg_rrm_fp_variant}
\BZ^{t+1} = \mathcal H\left(1_{\text{emb}} E^{\mathcal H}(\BX) + 1_{\text{prev}} \BZ^{t-r_{prev}},  \BZ^t \right),
\end{align}
for intermediate layers (with $t-r_{prev} \geq 0$), where $1_{\text{emb}},1_{\text{prev}} \in \{0,1\}$ 
denote whether the respective input is used (can be arbitrarily set and therefore define the design of the neural fixed-point architecture). 

\paragraph{Transformer-Style RRM Block.}
Following HRM/TRM-style designs, 
one RRM block consists of $L$ stacked Transformer-style layers. 
Let
\begin{align} \label{eq:bg_rrm_h0}
\BH_0 := 1_{\text{emb}} E^{\mathcal H}(\BX) + 1_{\text{prev}} \BZ^{t-r_{prev}} + \BZ^t.
\end{align}
For $l=0,\dots,L-1$, the update is
\begin{align} 
\BH'_{l} &= \text{Norm} \left(\BH_{l} + T^{D,I}(\BH_{l})\right), 
&& \underbar{D}\times \underbar{I}
\label{eq:bg_rrm_attn}\\
\BH_{l+1} &= \text{Norm}\!\left(\BH'_{l} + m^{D}(\BH'_{l})\right),
&& \underbar{D}\times I, \label{eq:bg_rrm_mlp}
\end{align}
and the block output is $\BZ^{t+1}=\BH_L$.
Here $T^{D,I}$ denotes self-attention along the position dimension and $m^D$ is a position-wise MLP (e.g., SwiGLU), while $\text{Norm}$ denotes RMSNorm, applied token-wise. In above equations, we tried to indicate the attention axes or axes used as features to neural operations with an underscore, while the neural operation operates independently on the other axes.

\paragraph{Output Mapping and Training with Deep Supervision.}
RRMs predict symbol distributions per position by mapping $\BZ^t$ to logits $\hat{\BY}^t\in\mathbb{R}^{I\times K}$ using an output head $\BO$ (typically linear in $D$), followed by a softmax over the $K$ symbols. 
Training uses deep supervision: losses (categorical cross-entropy) are applied to selected intermediate predictions, and gradients are truncated/detached across iteration segments.

\subsection{Limitation: Lack of Symbol Extrapolation in Vanilla RRMs}
\label{sec:background_symbol_limit}
Vanilla RRMs encode symbols via a learned embedding table $\Sigma\to\mathbb{R}^D$ inside $E^{\mathcal H}$. As a consequence, permuting symbols in the input can lead to differing results in the output. It also disallows the application to instances with unseen symbols $\Sigma'\supset \Sigma$ (i.e., $|\Sigma'|>K$), because new symbols require new embeddings. This is particularly restrictive for extrapolation across Sudoku sizes, where the alphabet grows with the grid (e.g., from digits $1:9$ to $1:16$ or $1:25$).

\paragraph{Data Augmentation.}
Previous methods combat this issue by augmenting the training data with symbol-permuted samples. While this can diminish performance inconsistencies, it requires additional computation due to the increased training data. Crucially, it still prevents extrapolation to unseen symbols.

\paragraph{Symbol-Axis Embedding.}
Conceptually, for each position $i$ and each symbol $c\in \Sigma$, SE-RRM stores whether $x_i=c$ along the symbol axis using a shared embedding vector for all usual symbols plus dedicated embeddings for special tokens such as mask/unknown, and zeros otherwise. 
This yields a tensor $E^{\mathcal C}(\BX)\in\mathbb{R}^{D\times I\times K}$ that can be augmented with positional and (optionally) task-type information,
$E^{\mathcal G} := E^{\mathcal C} + E^{\mathcal P} + E^{\mathcal T}.$

\paragraph{SE-RRM Fixed-Point Update.}
SE-RRM uses the same fixed-point structure but with a block
$\mathcal G:\mathbb{R}^{D\times I\times K}\times \mathbb{R}^{D\times I\times K}\to \mathbb{R}^{D\times I\times K},$ and

\begin{align} \label{eq:bg_serrm_fp}
\bar{\BZ}^{t+1} = \mathcal G \left(1_{\text{emb}} E^{\mathcal G}(\BX) + 1_{\text{prev}} \bar{\BZ}^{t-r_{prev}},\, \bar{\BZ}^t \right).
\end{align}

\paragraph{SE-RRM Block Architecture with Axial Attention over Positions and Symbols.}
Let $\BH_0$ be defined analogously to \eqref{eq:bg_rrm_h0}, but in $\mathbb{R}^{D\times I\times K}$. 
For $l=0,\dots,L-1$, SE-RRM applies two attention layers: first along positions, then along symbols, followed by a token-wise MLP:
\begin{align}
\BH'_{l} &= \text{Norm} \left(\BH_{l} + T^{D,I}(\BH_{l})\right),&& \underbar{D}\times \underbar{I}\times K\label{eq:bg_serrm_attn_pos}\\
\BH''_{l} &= \text{Norm} \left(\BH'_{l} + T^{D,K}(\BH'_{l})\right),&& \underbar{D}\times I\times \underbar{K}\label{eq:bg_serrm_attn_sym}\\
\BH_{l+1} &= \text{Norm} \left(\BH''_{l} + m^{D}(\BH''_{l})\right),\/&& \underbar{D}\times I\times K,\label{eq:bg_serrm_mlp}
\end{align}
and $\BZ^{t+1}=\BH_L$. 
Attention along the position and the MLP are uniform across the symbol axis and therefore do not break equivariance. Attention along the symbol dimension is itself permutation-equivariant. Therefore, equivariance is preserved, unless it is explicitly broken via task-type embeddings.

\paragraph{Output Head.}
SE-RRM projects each $D$-dimensional feature vector of $\BZ^{t}$ (for every position–symbol pair $(i,k)$) to a scalar logit, producing an output tensor in $\mathbb{R}^{1 \times I \times K} \cong \mathbb{R}^{I \times K}$. Categorical probabilities per position are then computed as in vanilla RRMs.




\begin{table}[htbp]
\centering
\caption{Notation used throughout the paper (RRMs/SE-RRMs, SAT solving, and G-RRM).\label{tab:notation}}
\resizebox{0.90\textwidth}{!}{%
\begin{tabular}{l c l}
\toprule
\textbf{Definition} & \textbf{Symbol} & \textbf{Type} \\
\midrule

\multicolumn{3}{l}{\textbf{Scalars and indices}} \\
grid dimension (Sudoku is $N\times N$) & $N$ & $\mathbb{N}$ \\
feature dimension & $D$ & $\mathbb{N}$ \\
number of positions/cells & $I$ & $\mathbb{N}$ \\
position / cell index & $i$ & $\{1,\ldots,I\}$ \\
number of symbols & $K$ & $\mathbb{N}$ \\
symbol (channel) index & $k$ & $\{1,\ldots,K\}$ \\
recurrent iteration index (RRM/SE-RRM) & $t$ & $\mathbb{N}$ \\
previous-state offset (recurrence hyperparameter) & $r_{\text{prev}}$ & $\mathbb{N}$ \\
number of layers per block & $L$ & $\mathbb{N}$ \\
layer index within a block & $l$ & $\{0,\ldots,L-1\}$ \\

\midrule
\multicolumn{3}{l}{\textbf{SAT}} \\

cell row, column, and value indices & $r,c,v$ & $\{1,\ldots,N\}$ \\
SAT variable: cell $(r,c)$ holds value $v$ & $x_{r,c,v}$ & $\{0,1\}$ \\
per-variable value ordering (arg sort of scores) & $\pi_i$ & permutation of $\{1,\ldots,K\}$ \\
value index / most-preferred value for variable $i$ & $d,\,d^{*}_i$ & $\{1,\ldots,K\}$ \\
second value index (At-Most-One clauses) & $w$ & $\{1,\ldots,N\}$ \\

\midrule
\multicolumn{3}{l}{\textbf{Sets and spaces}} \\
symbol alphabet (neural sections; may include mask) & $\Sigma$ & finite set, $|\Sigma|=K$ \\
extended symbol alphabet (unseen symbols) & $\Sigma'$ & $\Sigma'\supset\Sigma,\ |\Sigma'|>K$ \\
input/solution space (neural) & $\Sigma^I$ & tuples of symbols \\

\midrule
\multicolumn{3}{l}{\textbf{Tasks, solutions, and assignments}} \\
input task (neural encoding) & $\BX=(x_1,\ldots,x_I)$ & $\Sigma^I$ \\
solution task (neural target) & $\BY=(y_1,\ldots,y_I)$ & $\Sigma^I$ \\
symbol at position $i$ & $x_i,\,y_i$ & $\Sigma$ \\

\midrule
\multicolumn{3}{l}{\textbf{Embeddings and representations (neural)}} \\
RRM task embedding matrix & $E^{\mathcal H}(\BX)$ & $\Sigma^I \mapsto \mathbb{R}^{D\times I}$ \\
SE-RRM symbol-axis embedding tensor & $E^{\mathcal C}(\BX)$ & $\Sigma^I \mapsto \mathbb{R}^{D\times I\times K}$ \\
SE-RRM task embedding tensor & $E^{\mathcal G}(\BX)$ & $\Sigma^I \mapsto \mathbb{R}^{D\times I\times K}$ \\
positional embedding & $E^{\mathcal P}$ & $\mathbb{R}^{D\times I\times K}$ \\
task-type embedding & $E^{\mathcal T}$ & $\mathbb{R}^{D\times I\times K}$ \\
recurrent state at step $t$ (RRM) & $\BZ^t$ & $\mathbb{R}^{D\times I}$ \\
recurrent state at step $t$ (SE-RRM) & $\bar{\BZ}^t$ & $\mathbb{R}^{D\times I\times K}$ \\
learned initial state & $\BZ^0$ & $\mathbb{R}^{D\times I}$ \\
intermediate block states & $\BH_l,\BH'_l,\BH''_l$ & $\mathbb{R}^{D\times I}$ or $\mathbb{R}^{D\times I\times K}$ \\

\midrule
\multicolumn{3}{l}{\textbf{Models and blocks (neural)}} \\
RRM block (HRM/TRM) & $\mathcal H$ & $\mathbb{R}^{D\times I}\times\mathbb{R}^{D\times I}\to\mathbb{R}^{D\times I}$ \\
SE-RRM block & $\mathcal G$ & $\mathbb{R}^{D\times I\times K}\times\mathbb{R}^{D\times I\times K}\to\mathbb{R}^{D\times I\times K}$ \\
output head  & $\mathcal O$ & $\mathbb{R}^{D}\to\mathbb{R}^{K}$ \\
embedding inclusion indicator & $\mathbf{1}_{\text{emb}}$ & $\{0,1\}$ \\
previous-state inclusion indicator & $\mathbf{1}_{\text{prev}}$ & $\{0,1\}$ \\

\midrule
\multicolumn{3}{l}{\textbf{Neural operators}} \\
self-attention over positions & $T^{D,I}$ & $\mathbb{R}^{D\times I}\to\mathbb{R}^{D\times I}$ \\
self-attention over symbols & $T^{D,K}$ & $\mathbb{R}^{D\times K}\to\mathbb{R}^{D\times K}$ \\
MLP over features (SwiGLU) & $m^{D}$ & $\mathbb{R}^{D}\to\mathbb{R}^{D}$ \\
normalization (RMSNorm) & $\mathrm{Norm}$ & $\mathbb{R}^{D}\to\mathbb{R}^{D}$ \\

\midrule
\multicolumn{3}{l}{\textbf{Prediction}} \\
logits produced by the RRM backbone (single forward pass) & $\hat{\mathbf{Y}}$ & $\mathbb{R}^{I\times K}$ \\
logits at RRM iteration $t$ & $\hat{\BY}^{t}$ & $\mathbb{R}^{I\times K}$\\ 
score for assigning value $d$ to variable $i$ & $\hat{Y}_{i,d}$ & $\mathbb{R}$ \\

\bottomrule
\end{tabular}
}
\end{table}

\section{SE-RRM Model Training Details}
\label{app:trDetails}

\paragraph{Data.}
We use Sudoku puzzles of differing sizes to compare methods across problem specifications. For standard $9\times9$ Sudokus, we use the same data as \cite{freinschlag2026symbol}: 1,000 samples provided by \cite{wang2025hierarchical} to train the neural models and 1,000 test samples. The $16\times16$ and $25\times25$ data comprises 100 training samples for fine-tuning and 100 and 90 test samples, respectively, which are created with \textit{py-sudoku}. The average rate of empty cells differs per dataset, $9\times9$, $16\times16$, and $25\times25$ Sudokus have $68.88\%$, $63.49\%$, and $50.08\%$ empty cells, respectively.

\paragraph{Training Details.} 
We train an SE-RRM model initially on $9\times9$ Sudokus. We then fine-tuned it further for the $16\times16$ and $25\times25$ grids with 100 samples using a decreased learning rate and increased regularization.

\section{CP-SAT Experiments}
\label{app:cpsat}

To confirm findings from our main experiments, we provide additional experiments using the state-of-the-art constraint programming solver CP-SAT~\citep{perron2023cp}.

We compare four configurations designed to evaluate different levels of neural-symbolic coupling:

\begin{table}[htbp]
\centering
\caption{Tested CP-SAT solver configurations}
\label{tab:cpsat_configurations}

\begin{threeparttable}
\setlength{\tabcolsep}{4pt}
\renewcommand{\arraystretch}{1.15}

\resizebox{0.7\linewidth}{!}{%
\begin{tabular}{lcc}
\toprule
ID & Name & Description \\
\midrule
C1 & Default CP-SAT & Full CP-SAT heuristics \\
C2 & Naive Fixed & Branches on most constrained variable and chooses lowest digits first \\
C3 & Soft Hint & Warm-start with SE-RRM \\
C4 & Repaired Hint & Propagates SE-RRM assignment as partial solution and resolves conflicts \\
\bottomrule
\end{tabular}
}

\end{threeparttable}
\end{table}

All four configurations share the same underlying CP-SAT base model, enforcing $N \times N$ Sudoku constraints (AllDifferent constraints across all rows, columns, and subgrids). To isolate the impact of search guidance and ensure reproducibility, all configurations are restricted to a single worker thread and a fixed random seed.

The configurations are defined as follows:
\begin{itemize}
    \item \textbf{C1 (Default CP-SAT)} functions as the primary symbolic baseline, utilizing the solver's default configuration.
    \item \textbf{C2 (Naive Fixed)} represents a blind symbolic baseline. It enforces a fixed search strategy that branches on the variable with the minimum domain size and selects the minimum value first. This matches the branching heuristics of the naive backtracking analyzed in Section~\ref{sec:backtracking}, while retaining CP-SAT's propagation and conflict-driven clause learning.
    \item \textbf{C3 (Soft Hint)} warm-starts CP-SAT with the SE-RRM's argmax predictions via \texttt{model.AddHint}. If the hints are globally consistent, the solver completes in the hint phase with zero conflicts; when a conflict arises, it continues following the hints for up to the default budget (\texttt{hint\_conflict\_limit = 10}), using each conflict to learn which hint values are wrong, before discarding the hints and falling back to automatic search.
    \item \textbf{C4 (Repaired Hint)} uses the same hints with  \texttt{repair\_hint = True}: rather than discarding on conflict, 
the solver runs a propagation-based local search that adjusts only the inconsistent cells while preserving the rest. A successful repair yields zero conflict-driven clause learning backtracking; a larger budget (\texttt{hint\_conflict\_limit = 100}) funds this process.
\end{itemize}

The performance of these configurations across different grid sizes is presented in Table~\ref{tab:cpsat_results}.

\begin{table}[htbp]
\centering
\caption{CP-SAT performance across different constraint configurations. Lower values are better.}
\label{tab:cpsat_results}
\setlength{\tabcolsep}{9pt}
\renewcommand{\arraystretch}{1.12}

\begin{tabular}{lccc}
\toprule
Configuration & $9\times9$ & $16\times16$ & $25\times25$ \\
\midrule
\multicolumn{4}{c}{\emph{Mean search conflicts}} \\
\addlinespace[1pt]
C1 & 21.8 & 32.8 & 5.6 \\
C2 & 28.1 & 35.8 & 3.9 \\
C3 & 4.7  & 27.6 & 6.0 \\
C4 & \textbf{0.0} & \textbf{16.4} & \textbf{3.8} \\
\addlinespace[4pt]
\multicolumn{4}{c}{\emph{Mean wall-clock time (s)}} \\
\addlinespace[1pt]
C1 & \textbf{0.0064} & \textbf{0.0237} & \textbf{0.0298} \\
C2 & 0.0122 & 0.0366 & 0.0491 \\
C3 & 0.0066 & 0.0312 & 0.0409 \\
C4 & 0.0125 & 0.0971 & 0.0469 \\
\bottomrule
\end{tabular}
\end{table}

The results are best understood through \emph{hint feasibility}: during presolve, CP-SAT classifies the SE-RRM hint set as either feasible---the predictions form a globally consistent assignment, allowing the solver to complete in the hint phase with zero branching---or infeasible, requiring repair or general search. On feasible instances, C3 and C4 are equivalent; they diverge only on infeasible ones.

\begin{enumerate}
    \item \textbf{Conflict Reduction at Low Density:} For $9\times9$ grids (clue density $\approx 31.3\%$), conflict counts are right-skewed: propagation resolves most instances, while a hard tail drives the mean. C2's $28.1$ mean conflicts already reflect CP-SAT's full conflict-driven clause learning---C2 is a blind \emph{branching heuristic}, not a blind solver. C4 achieves $0.0$ mean conflicts: the $91.1\%$ feasible instances complete in the hint phase, and all remaining infeasible instances are fully repaired within the 100-conflict budget. C3 reduces mean conflicts by $83\%$ ($4.7$ vs.\ $28.1$): on infeasible instances, it follows wrong hints for up to 10 conflicts---building clause knowledge from each failure---before handing off to conflict-driven clause learning.

    \item \textbf{Propagation-Dominated Regimes at Scale:} At $25\times25$ (clue density $\approx 49.9\%$), propagation resolves $92.2\%$ of instances under C2 with zero conflicts, leaving little room for any branching heuristic to influence outcomes. Mean conflict counts collapse across all configurations. Notably, C3 performs worse than the blind C2 baseline ($6.0$ vs.\ $3.9$): incorrect hints add overhead to otherwise near-trivial solves. C4 beats C2 by a negligible margin ($3.8$ vs.\ $3.9$). This illustrates a fundamental limit of hint-based guidance: when propagation already resolves most of the search space, the branching heuristic has little opportunity to contribute, and incorrect hints are a net cost.

    \item \textbf{Wall-Clock Time and Overhead:} C3 consistently reduces wall-clock time relative to C2: $1.9\times$ on $9\times9$ ($0.0066$\,s vs.\ $0.0122$\,s), $1.2\times$ on $16\times16$ ($0.0312$\,s vs.\ $0.0366$\,s), and $1.2\times$ on $25\times25$ ($0.0409$\,s vs.\ $0.0491$\,s). C1's automatic portfolio heuristics remain faster than C3 at all scales. C4 minimizes conflicts across all grid sizes but incurs substantial overhead at $16\times16$ ($0.0971$\,s vs.\ $0.0237$\,s for C1): with only $22.0\%$ of hint sets feasible, the repair budget is frequently exhausted before a consistent assignment is found, forcing costly fallback to general search.
\end{enumerate}

\end{document}